\documentclass[letterpaper]{article} 
\usepackage{aaai2026}  
\nocopyright                
\usepackage{times}  
\usepackage{helvet}  
\usepackage{courier}  
\usepackage[hyphens]{url}  
\usepackage{graphicx} 
\usepackage{natbib}  
\usepackage{caption} 
\usepackage{algorithm}
\usepackage{algorithmic}
\usepackage{booktabs} 
\usepackage{multirow} 
\usepackage{siunitx}  
\usepackage{tabularx}

\usepackage{subcaption}
\usepackage{cuted}
\usepackage{adjustbox}
\usepackage{array}
\usepackage{newfloat}
\usepackage{listings}
\DeclareCaptionStyle{ruled}{labelfont=normalfont,labelsep=colon,strut=off} 
\floatstyle{ruled}
\newfloat{listing}{tb}{lst}{}
\floatname{listing}{Listing}
\title{From Diagnosis to Improvement: Probing Spatio-Physical Reasoning \\ in Vision Language Models}

\author{Tiancheng Han\textsuperscript{\rm 1, \rm 2},
Yunfei Gao\textsuperscript{\rm 3},
Yong Li\textsuperscript{\rm 1},
Wuzhou Yu\textsuperscript{\rm 1},
Qiaosheng Zhang$^{\dagger}$\textsuperscript{\rm 2,\rm 4},
\begingroup
\renewcommand\thefootnote{\textdagger}
Wenqi Shao$^{\dagger}$\textsuperscript{\rm2, \rm4}
\endgroup
}

\affiliations{
\textsuperscript{\rm 1}Tongji University,
\textsuperscript{\rm 2}Shanghai Innovation Institute,
\textsuperscript{\rm 3}East China University of Science and Technology,
\textsuperscript{\rm 4}Shanghai AI Laboratory
}

\usepackage{bibentry}

\begin{document}

\maketitle

\begin{abstract}
Spatio-physical reasoning, a foundation capability for understanding the real physics world, is a critical step towards building robust world models. While recent vision language models (VLMs) have shown remarkable progress in specialized domains like multimodal mathematics and pure spatial understanding, their capability for spatio-physical reasoning remains largely unexplored. This paper provides a comprehensive diagnostic analysis of mainstream VLMs, revealing that current models perform inadequately on this crucial task. Further detailed analysis shows that this underperformance is largely attributable to biases caused by human-like prior and a lack of deep reasoning. To address these challenges, we apply supervised fine-tuning followed by rule-based reinforcement learning to Qwen2.5-VL-7B, resulting in significant improvements in spatio-physical reasoning capabilities and surpassing leading proprietary models. Nevertheless, despite this success, the model's generalization to new physics scenarios remains limited---underscoring the pressing need for new approaches in spatio-physical reasoning.
\end{abstract}



\section{Introduction}

\begingroup
\renewcommand\thefootnote{} 
\footnotetext{
    \textsuperscript{\textdagger}Corresponding author: zhangqiaosheng@pjlab.org.cn, \\
    \phantom{\textsuperscript{\textdagger}1231231231231231231231}shaowenqi@pjlab.org.cn
}
\endgroup

The ability to intuitively reason about the physical world—predicting how objects interact under physical laws—is a cornerstone of intelligence. This capability, which we term \textbf{Spatio-Physical Reasoning}, involves both perceiving spatial information and reasoning with physical laws based on that perception. It is a fundamental prerequisite for any agent to operate effectively in the real world.

Recent advances in Vision Language Models (VLMs) have shown remarkable capabilities in various tasks. However, existing research on their reasoning abilities has been largely confined to specific domains like multimodal mathematics \cite{meng2025mmeureka} or pure spatial understanding \cite{ouyang2025spacer,liao2025improvedvisualspatialreasoningr1zerolike}, which involves only spatial relations and does not require any physical reasoning. When physics is involved, it is often in well-defined, textbook-style contexts \cite{shen2025phyxdoesmodelwits,xiang2025seephysdoesseeinghelp,dai2025physicsarenamultimodalphysicsreasoning,zheng2025scalingphysicalreasoningphysics}, leaving a critical gap in understanding VLMs' performance in spatio-physical reasoning tasks.

Bridging this gap is essential for ambitious AI goals like World Models, as a robust internal model of physical outcomes is a potential pathway toward AGI \cite{assran2025vjepa2selfsupervisedvideo}. Furthermore, strong physical cognition directly improves performance in downstream embodied AI tasks \cite{chow2025physbench}. In this work, we first conduct a diagnostic analysis of how mainstream VLMs behave in spatio-physical reasoning, and then investigate whether fine-tuning can remedy their deficiencies and improve generalization.

Our main contributions are: (1) We conduct a comprehensive diagnostic analysis of VLM's underperformance in spatio-physical reasoning. Moving beyond simple accuracy metrics, we identify the root causes of their poor performance: systematic, human-like biases and a lack of deep reasoning. Crucially, our analysis reveals that the reasoning quality, not the mere presence of the reasoning process, is the decisive factor for success. (2) We systematically examine the efficacy and generalization abilities of fine-tuning. We show that a combination of supervised fine-tuning (SFT) and reinforcement learning (RL) can significantly boost VLM's in-domain performance in spatio-physical reasoning tasks, and the resulting model even outperforms leading proprietary models. On the other hand, its ability to foster generalization to novel physical contexts remains limited. This finding highlights the challenge for current paradigms in moving beyond pattern matching to instill robust, generalizable physical principles.

\begin{table*}[t]
\centering
\small
\begin{tabularx}{\textwidth}{l l *{10}{>{\raggedleft\arraybackslash}X}}
\toprule
\multirow{2}{*}{\textbf{Model}} & \multirow{2}{*}{\textbf{Accuracy}} & \multicolumn{2}{c}{\textbf{Difficulty Bias}} & \multicolumn{5}{c}{\textbf{Height Bias}} & \multicolumn{3}{c}{\textbf{Duplicated Height Bias}} \\
\cmidrule(lr){3-4} \cmidrule(lr){5-9} \cmidrule(lr){10-12} 
& & Easy & Hard & 2 & 3 & 4 & 5 & 6 & 2 & 4 & 6 \\
\midrule
InternVL3-8B        & 0.542 &0.794& 0.385 & 0.938 & 0.805 & 0.472 & 0.221 & 0.249  &0.996 &0.700 &0.252 \\
InternVL3-14B        & 0.547 & 0.711 & 0.405 & 0.984 & 0.819 & 0.505 & 0.013 & -0.162 &0.988 &0.622 &-0.354 \\
InternVL3-38B        & 0.547 & 0.997 & 0.967 & 0.999 & 0.993 & 0.978 & 0.942 & 0.868 &1.000 &0.939 &0.804 \\
InternVL3-78B        & 0.575 & 0.430 & -0.311 & 0.764 & 0.264 & -0.144 & -0.317 & -0.421 &1.000 &0.798 &0.076  \\
Qwen2.5VL-7B        & 0.522 & 0.218 & -0.115 & 0.059 & 0.281 & -0.107 & 0.118 & -0.094 &-0.032 &0.038 &-0.179 \\
Qwen2.5VL-32B       & 0.546 & 0.927 & 0.335 & 0.938 & 0.692 & 0.692 & 0.573 & 0.317 &0.258 &0.514 &-0.036 \\
Qwen2.5VL-72B       & 0.547 & 0.967 & 0.596 & 0.724 & 0.781 & 0.933 & 0.928 & 0.715 &0.018 &1.0 &0.722 \\
Gemma3-12B-it           & 0.507 & -0.698 & -0.714 & -0.674 & -0.681 & -0.713 & -0.735 & -0.736 &-0.653 &-0.695 &-0.748 \\
Gemma3-27B-it           & 0.550 & -0.533 & -0.663 & -0.297 & -0.582 & -0.640 & -0.705 & -0.733 &-0.145 &-0.524 &-0.736 \\
\midrule
GPT-4o               & 0.560 & - & - & - & - & - & - & -& - & - & - \\
o4-mini          & 0.593 & - & - & - & - & - & - & - & - & - & -\\
o3        & 0.641 & - & - & - & - & - & - & - & - & - & -\\
\midrule
Human Expert & 0.779 & - & - & - & - & - & - & - & - & - & -\\
\bottomrule
\end{tabularx}
\caption{Accuracy and behavioral modeling parameters. $T_{\text{pref}}$ is used to quantify models' bias. A positive $T_{\text{pref}}$ indicates the model tends to answer true, while a negative score indicates a preference for false.}
\label{tbleval}
\end{table*}

\section{Related Works}
\paragraph{Enhancing Reasoning Ability in VLMs}
Recent efforts to enhance reasoning in VLMs have focused on specialized domains. In multimodal mathematical reasoning, approaches such as two-stage reinforcement learning~\cite{peng2025lmmr1empowering3blmms} and online filtering~\cite{meng2025mmeureka} have achieved notable success. In parallel, spatial reasoning has been advanced through large-scale video or 3D datasets and tailored training schemes to improve understanding of spatial layouts~\cite{ouyang2025spacer, daxberger2025mmspatialexploring3dspatial}.


\paragraph{Physical Reasoning}
Research in physical reasoning follows two main directions: evaluation and enhancement. Early evaluation benchmarks relied on physics simulators to test intuitive dynamics and statics (e.g., IntPhys \cite{riochet2020intphysframeworkbenchmarkvisual}, ShapeStacks \cite{groth2018shapestacks}, Physion \cite{physion}), while recent benchmarks assess symbolic knowledge \cite{zhang2025physreasoncomprehensivebenchmarkphysicsbased} or real-world physical understanding \cite{chow2025physbench}. In contrast, our work targets a more structured and physical fine-grained task: static stability analysis. Enhancement efforts either integrate external modules for contextual scene modeling \cite{balazadeh2025physicscontextbuildersmodular} or pursue end-to-end learning of general physical commonsense \cite{nvidia2025cosmosreason1physicalcommonsense}, which may not prioritize fine-grained physical laws.


\section{Modeling the Reasoning Behavior of VLMs}

We analyze VLM behaviors on a prototypical spatio-physical reasoning task: judging the stability of stacked objects using the ShapeStacks benchmark (Groth et al., 2018) generated by MuJoCo engine \cite{todorov2012mujoco}. This benchmark features stacked objects of varying heights (2-6) and difficulty levels, requires models to reason by combining the spatial information of the stacked system with the centre of mass criterion. Models classify scenes as balanced (`True') or unbalanced/will collapse (`False') based on multiple viewpoints. Our analysis, conducted on an 888-sample test subset, examines model performance, reasoning strategies, error types, and inherent biases. Further details about the dataset are provided in Appendix A.

\begin{figure}[h!]
    \centering
    \includegraphics{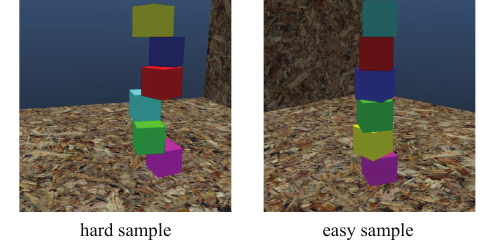} 
    \caption{Easy and hard samples in the ShapeStacks benchmark. Although both examples shown are in equilibrium, the hard sample exhibits noticeable misalignment between different layers, which can easily lead to misjudgment.}
    \label{figdifficulty}
\end{figure}

\subsection{Evaluation}
We evaluate 9 mainstream open-source VLMs ($\geq$7B), including their instruction-tuned variants except for the InternVL series\footnote{InternVL3 Instruct models were not trained with mixed preference optimization \cite{zhu2025internvl3}}, along with several proprietary models (e.g., GPT-4o, o3, o4-mini) for reference. For open-source models, we evaluate three times and report the average; for proprietary models, we evaluate only once. Hyperparameter settings are provided in Appendix B.1. As shown in Table~\ref{tbleval}, none of the open-source models achieve accuracy above 0.6, and even the leading proprietary model, o3, reaches only 0.641—well below the human expert performance of 0.779. Human evaluation details are provided in Appendix B.2.

\begin{figure*}[t]
    \centering
    \includegraphics{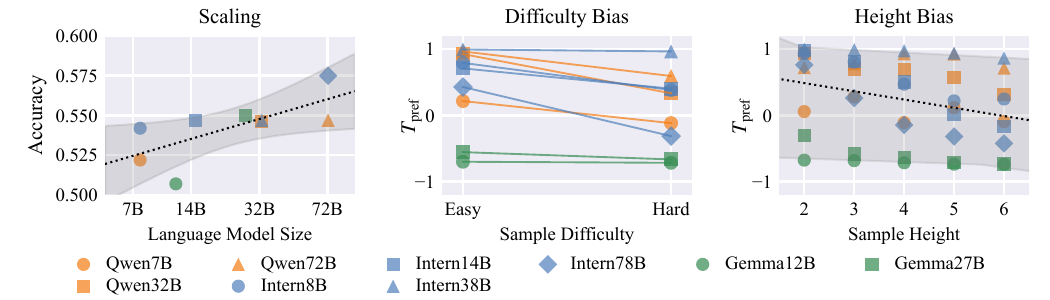}
    \caption{Left: Model accuracy exhibits a positive correlation with the log size of the language model component. Center: Model bias ($T_{\text{pref}}$) shifts towards predicting ``False'' on hard samples. Right: Models become increasingly biased towards ``False'' as stack height increases. In the left plot, trend line and 95\% CI grey shadow are fitted using linear regression. In the right plot, the shaded envelope of trend lines for each model and dash line for overall trend is fitted with linear mixed-effect model.}
    \label{figeval}
\end{figure*}

Further, we observe a weak but statistically significant positive correlation between accuracy and the logarithm size of the language model of VLMs (exclude the vision encoder) as shown in Figure \ref{figeval}, left. A linear regression model confirms this trend, yielding a fixed-effect slope of 0.0352 (95\% confidence interval $[0.0031,0.0673]$) with $p=0.0360$. This shows that the scaling law still holds for spatio-physical reasoning tasks and that performance is more closely associated with the size of the language model, as VLMs with different overall sizes may share vision encoders of the same size. Full details are provided in Appendix B.4.

\subsection{Cognitive Behavior and Error Cause}
To understand the underlying causes of these underperformance, we conduct a manual analysis of their Chain-of-Thought (CoT) responses to characterize their logical steps, identify common error sources, and examine the role of higher-level cognitive behaviors.

\subsubsection{Step-by-Step Inspection of CoT}
Through manual analysis of the model responses, we identify a stepwise behavioral pattern in the model's CoT: \emph{(1) Task Identification}: The model first interprets the task definition presented in the prompt. 
\emph{(2) Visual Analysis}: The model then analyzes the visual information in the image as a basis for reasoning. 
\emph{(3) Physical Reasoning}: Based on the task goal and visual input, the model applies relevant physical principles to perform reasoning. 
\emph{(4) Conclusion}: Finally, the model synthesizes the above information to reach a final decision.

Notably, the model often adopts a process of elimination approach: checking whether any object violates stability conditions and concluding stability only if none is found.

\subsubsection{Three Main Error Causes}
We identify three primary sources of failure: \textbf{(1) Visual Perception Errors}: While the model rarely misjudges the number, color, or approximate location of objects in a scene, it frequently makes errors in identifying fine-grained spatial relationships like layer misalignment. This suggests that the model is only capable of recognizing relatively salient visual cues. \textbf{(2) Physical Reasoning Errors}: the model applies incorrect physical principles. For example, treating a horizontally lying cylinder as if it were upright when estimating center of mass. \textbf{(3) Causal Reasoning Errors}: These errors arise when the model draws incorrect conclusions from otherwise valid intermediate inferences. A common failure mode is the model treating visual stillness as evidence of stability (i.e., ``the system hasn’t moved, so it must be stable''), despite the prompt explicitly stating that the images show a static moment.

Among these, visual perception errors are the most critical: as revealed by our earlier analysis of CoT, an incorrect visual input inevitably leads to following incorrect reasoning and predictions. Causal reasoning errors are largely related to the model’s logical priors, rather than spatial perception or physical priors.

\subsubsection{Rich High-level Cognitive Behaviors with Limited Utility}
Inspired by previous research~\cite{gandhi2025cognitivebehaviorsenableselfimproving}, which identifies high-level cognitive behaviors as key to performance in symbolic reasoning, we investigate whether these same behaviors can also boost VLMs' performance in spatio-physical reasoning. The four cognitive behaviors under consideration are: \textbf{Verification}: checking or comparing a generated result to confirm its correctness. \textbf{Backtracking}: abandoning a failed reasoning path and explicitly attempting a new strategy. \textbf{Subgoal Setting}: decomposing a complex problem into a sequence of smaller, more manageable subproblems. \textbf{Backward Chaining}: reasoning backward from the desired goal to infer the necessary preconditions to achieve it. 

Based on this, we investigate whether the presence of such behaviors correlates with correctness in our task. We use the o4-mini model to annotate the responses of the InternVL series, comparing the distribution of these behaviors in correct and incorrect responses. More details are presented in Appendix C.

As shown in Figure \ref{figcog}, for all InternVL models, the distribution of cognitive behaviors does not significantly differ between correct and incorrect responses. This suggests that the quality and content of the reasoning process matter more than the mere presence of cognitive behavior patterns. The model is simply performing superficial rather than genuinely deep reasoning.


\begin{figure}[h!]
    \centering
    \includegraphics{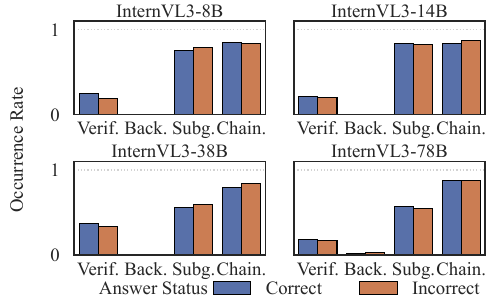} 
    \caption{The proportion of advanced reasoning behaviors in correct and incorrect responses of the InternVL model series shows no significant difference.}
    \label{figcog}
\end{figure}

\begin{figure*}[t]
    \centering
    \begin{subfigure}[b]{1.0\columnwidth}
        \centering
\includegraphics{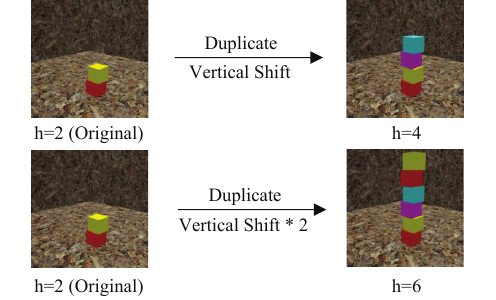}
        \caption{}
        \label{fig3a}
    \end{subfigure}
    \begin{subfigure}[b]{1.0\columnwidth}
        \centering
\includegraphics{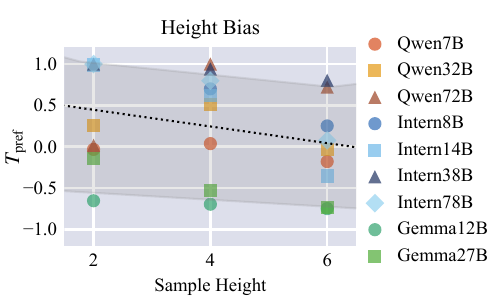}
        \caption{}
        \label{fig3b}
    \end{subfigure}
    \caption{(a) Generate higher samples with highly similar mechanical structures through duplication and translation. (b) Bias still holds for most models on duplicated samples. Fitting with a linear mixed-effects model reveals a general decreasing trend as height increases. Although certain models show an increasing tendency to predict ``true'' with greater height, these deviations do not affect the overall trend. The grey shadow is the envelope of trend lines for each model.}
    \label{fig3}
\end{figure*}

\subsection{Bias with Human-like Prior}

We next investigate if VLMs, pre-trained on human-generated data, exhibit human-like biases in spatio-physical reasoning. Specifically, we consider two priors: the \emph{Difficulty Prior}, where greater inter-object displacement are considered as instability, and the \emph{Height Prior}, where taller stacks are perceived as less stable.

To quantify this, we analyze the confusion matrix of model predictions and define the preference score $T_{\text{pref}}$:

\begin{equation}
\label{tpref}
T_{\text{pref}}:=\tanh(\psi),\quad\psi := \frac{\text{Recall} - \text{Specificity}}{\text{Specificity}}
\end{equation}
where $\text{Recall} := \frac{\text{TP}}{\text{TP} + \text{FN}}$ and $\text{Specificity} := \frac{\text{TN}}{\text{TN} + \text{FP}}$ indicate the proportion of correctly predicted True and False samples, respectively. The metric $T_{\text{pref}}$ reflects the model's preference: a positive score indicates a bias toward “True”, while a negative score indicates a bias toward “False”.

\subsubsection{Difficulty Bias with Inter-object Displacement Prior}
For the inter-object displacement prior, we analyze the model's preference score $T_{\text{pref}}$ on easy and hard samples. As shown in Figure~\ref{figdifficulty}, hard samples involve large misalignment in both stable and unstable cases, which can mislead the model. We hypothesize that if the model has a human-like bias, it will assign lower $T_{\text{pref}}$ scores to hard samples, mistakenly judging balanced stacks as unbalanced.

Our hypothesis is supported by the analysis of the response data used in the previous evaluation. As shown in the “Difficulty Bias” column of Table~\ref{tbleval} and the middle plot of Figure \ref{figeval}, all models yield lower $T_{\text{pref}}$ score on hard samples compared to easy ones. This confirms that the visual displacement features in hard samples can be misleading to the models. The consistent bias pattern suggests that the models exhibit a prior similar to humans: noticeable inter-object displacement is associated with instability.

Additionally, the model's ``elimination-based'' reasoning process may also lead to this bias: in hard samples, large inter-layer misalignments are often identified as instability factors, which then directly leads models to conclude that the structure is unbalanced.

\subsubsection{Height Bias with Height Prior}
To investigate the height prior, we first analyze the $T_{\text{pref}}$ score across stack heights. As shown in the Height Bias column of Table~\ref{tbleval} and the right panel of Figure \ref{figeval}, models tend to exhibit lower $T_{\text{pref}}$ scores as height increases. A linear mixed-effects model analysis further confirms this trend, yielding a fixed-effect slope of $-0.1244$ (95\% confidence interval: $[-0.2025, -0.0463]$, $p = 0.0018$), indicating statistical significance. 

To test the robustness of this bias, we create duplicated samples: based on $h=2$ cube samples, we generate $h=4$ and $h=6$ versions by vertically stacking identical cubes without altering horizontal positions (Figure \ref{fig3a}). Although these samples differ in height, their mechanical structure remains essentially the same. 

On these duplicated samples, most models—especially the InternVL and Gemma series—exhibit a decreasing trend in $T_{\text{pref}}$ as object height increases (Figure~\ref{fig3b}, Table~\ref{tbleval}). This suggests a persistent height-related bias. A linear mixed-effects model yields a fixed-effect slope of $-0.1008$ (95\% confidence interval: $[-0.1965, -0.0051]$, $p = 0.0389$), indicating a statistically significant negative correlation and thus a general tendency toward this bias. The Qwen series' bias, however, disappears on these samples, demonstrating that its prior is brittle and less systematic. Details for fitting are provided in Appendix B.4.

These two experiments confirm that a human-like height prior is a common VLM feature, though its strength varies across model families. We speculate that this height bias is influenced by real-world data distributions, where higher objects are often less stable than lower ones. Models trained on such data are likely to inherit this bias. Furthermore, the varying strength of this bias across different VLMs may be attributed to their distinct training methodologies.

\begin{figure*}[t]
    \centering
    \begin{subfigure}[b]{0.588\linewidth}
        \centering
        \includegraphics{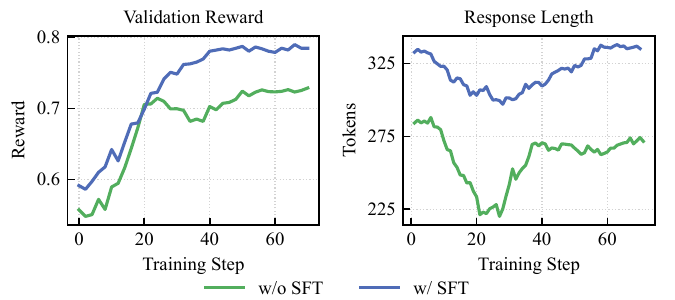}
        \caption{}
        \label{figtrain_a}
    \end{subfigure}
    \begin{subfigure}[b]{0.392\linewidth}
        \centering
        \includegraphics{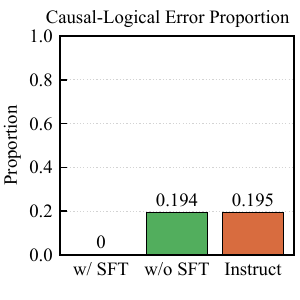}
        \caption{}
        \label{figtrain_b}
    \end{subfigure}
    
    \caption{The benefits of SFT as an initialization for RL. We compare RL fine-tuning with SFT initialization (w/ SFT) against RL fine-tuning from the Instruct model (w/o SFT). (a) SFT initialization leads to more stable validation rewards and maintains longer responses during training. (b) SFT also eliminates the causal-logical errors that are present without it.}
    \label{figtrain}
\end{figure*}

\subsection{Transfer of Prior Knowledge}
To investigate prior knowledge transfer, we evaluate two specialized models: SpaceR \cite{ouyang2025spacer} with a strong spatial prior, and MM-Eureka-Qwen-7B \cite{meng2025mmeureka} with a strong physical prior. We compare their performance against their base model, Qwen2.5-VL-7B on ShapeStacks.

As shown in Table~\ref{tbleureka}, despite their strong priors, these models exhibit almost no accuracy improvement, indicating no performance gain. This suggests that fine-tuning on a single domain is insufficient for compositional reasoning.

\begin{table}[h]
\small
\centering
\begin{tabular}{lc}
\toprule
\textbf{Model} & \textbf{Accuracy} \\
\midrule
Qwen2.5-VL-7B      & 0.522          \\
MM-Eureka-Qwen-7B         & 0.521             \\
SpaceR          & 0.522          \\
\bottomrule
\end{tabular}
\caption{Prior Knowledge Transfer: Possessing spatial or physical reasoning abilities alone makes it difficult to improve performance in spatio-physical reasoning tasks.}
\label{tbleureka}
\end{table}


\section{Improvement through Fine-Tuning}
Building on our diagnostic analysis, this section investigates whether fine-tuning can remedy the identified VLM limitations. We show that a two-stage training pipeline--- first supervised fine-tuning (SFT) and then reinforcement learning (RL)---on Qwen2.5-VL-7B significantly improves its performance on ShapeStacks, and achieves state-of-the-art performance compared to leading proprietary models. Furthermore, we test the improved model's generalization to novel physical scenarios across dimensions, dynamics, and heights. We find that the model exhibits a certain degree of generalization, but its performance degrades as the data distribution shifts away from the training domain.

\begin{figure*}[t]
    \centering
    \includegraphics{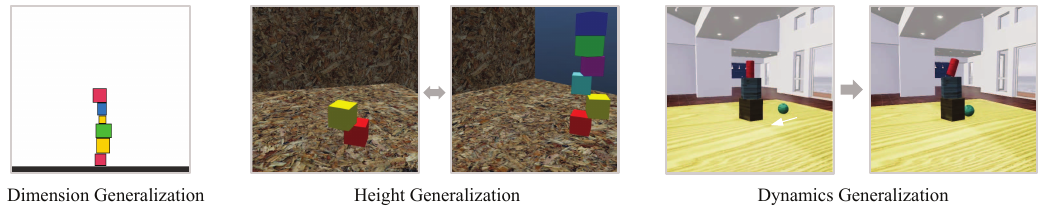}
    \caption{An illustration of the three physics-based generalization tests. Dimensional generalization tests transfer between 2D and 3D data. Height generalization tests transfer between low and high stacks. Dynamics generalization tests transfer from static stability to dynamic scenarios involving external forces (Physion).}
    \label{fig_generalization}
\end{figure*}

\begin{table*}[t]
\centering
\small 
\newcolumntype{P}[1]{>{\centering\arraybackslash}p{#1}}
\begin{tabular}{l lllll}
\toprule
\multirow{2}{*}{\textbf{Models}} & \multicolumn{5}{c}{\textbf{Test Samples}} \\
\cmidrule(l){2-6} 
& ShapeStacks & ShapeStacks high & ShapeStacks low & 2d-towers & Physion Support \\
\midrule
Qwen2.5VL-7B & 0.522 & 0.580 & 0.531 & 0.510 & 0.504 \\

Qwen-SFT-RL (ShapeStacks) & \textbf{0.771 / 47.7\%} & \textbf{0.781 / 34.7\%} & \textbf{0.799 / 50.5\%} & 0.658 / 29.0\% & 0.509 / 1.0\% \\

Qwen-high (ShapeStacks high) & 0.606 / 16.1\% & \textbf{0.678 / 16.9\%} &0.573 / 7.9\% &0.670 / 31.4\% & 0.493 / -2.2\%\\

Qwen-low (ShapeStacks low) & 0.677 / 29.7\% & 0.683 / 17.8\% & \textbf{0.751 / 41.4\%} & 0.661 / 29.6\%& 0.507 / 0.6\% \\

Qwen-2d (2d-towers) & 0.613 / 17.4\% & 0.674 / 16.2\% & 0.609 / 14.7\% & \textbf{0.918 / 80.0\%} & 0.522 / 3.6\% \\

\midrule

Qwen-RL (ShapeStacks) & \textbf{0.745 / 42.7\%} & \textbf{0.766 / 32.1\%} & \textbf{0.792 / 49.2\%} & 0.520 / 2.0\% & 0.5 / -0.8\% \\
Qwen-SFT (ShapeStacks) & \textbf{0.524 / 0.4\%} & \textbf{0.532 / -8.3\%} & \textbf{0.521 / -1.9\%} & 0.590 / 15.7\% & 0.493 / -2.2\% \\

\bottomrule
\end{tabular}
\caption{Metrics in Generalization Ability Analysis. In the ``Model'' column, the dataset used for fine-tuning is indicated in brackets after the model name.The fine-tuned model evaluation results are presented in the format: ``accuracy / improvement over the pre-fine-tuned Instruct model''. Bolded numbers indicate in-domain evaluation results.}
\label{tblgeneralization}
\end{table*}

\subsection{In-domain Performance Improvement}
\subsubsection{Experimental Setup}
Our fine-tuning data is the ShapeStacks training set (13,618 samples). We construct the SFT dataset by using the o4-mini model to distill CoT-style reasoning responses for 3,000 randomly selected samples (seed=0). The remaining samples are formatted for reinforcement learning with outcome supervision.

\subsubsection{Training}
We fine-tune Qwen2.5-VL-7B using the TRL framework \cite{vonwerra2022trl} for SFT and veRL \cite{sheng2024hybridflow} with the GRPO algorithm for RL. For SFT, we train for 40 epochs using LoRA with a learning rate of 1e-4, which is decayed using a cosine scheduler with 5\% warmup. For RL stage, we perform full-parameter fine-tuning for 8 epochs. The learning rate is set to 2e-6 (with a cosine scheduler and 10\% warmup) and a batch size of 1024. For each prompt, we rollout 6 responses, with the total reward being a weighted sum of a format score and a correctness score. We require the model to output the CoT and the final answer between different special tokens (placing the CoT between `\textless think\textgreater' and `\textless /think\textgreater', and the final answer between `\textless answer\textgreater' and `\textless/answer\textgreater'), which is used as a binary format reward (0 or 1). The correctness of the model's final answer is also used as a binary answer reward. In the total reward, the format reward accounts for 10\%, and the answer reward accounts for 90\%. The resulting models are named as Qwen-SFT-RL. Details are provided in Appendix D.1.

\subsubsection{Evaluation}
We evaluate the Qwen-SFT-RL model on the ShapeStacks test set used in the previous section. The results are shown in Table~\ref{tblgeneralization}. As the table indicates, SFT+RL significantly improves the model's performance, achieving a 47.7\% gain over the Instruct model and outperforming the leading proprietary model o3. This demonstrates that the SFT+RL fine-tuning paradigm can effectively enhance the performance of VLMs on stacked-system stability tasks.

\subsubsection{Ablation Study on SFT}
We conduct an ablation study to understand SFT's role. We create a Qwen-RL model by applying RL directly to the base model without the SFT stage. We also evaluate the Qwen-SFT model, which is fine-tuned using SFT only. The hyperparameters are the same as those used for training Qwen-SFT-RL.


As shown in Table \ref{tblgeneralization}, Qwen-RL achieves substantial 42.7\% accuracy improvement but remains below Qwen-SFT-RL's 47.7\%, indicating RL alone is suboptimal. While Qwen-SFT shows minimal improvement (0.4\%), training dynamics analysis (Figure~\ref{figtrain}) reveals SFT initialization benefits. Qwen-RL exhibits overfitting (validation reward drops sharply around step 20), fails to maintain response length, and retains the base model's causal reasoning error rate, unlike error-free Qwen-SFT-RL.

This suggests that RL with outcome supervision alone is insufficient to guide reasoning. SFT serves two critical roles: (1) it provides a robust initialization for RL, enhancing training stability and final performance, and (2) its CoT data constrains the model to learn correct reasoning logic.

\subsection{Generalization Ability Test}
We investigate whether the model can generalize specific spatio-physical reasoning patterns through post-training. Specifically, we explore three types of generalization based on different feature patterns.
\begin{itemize}
    \item \textbf{Dimensional generalization}: We evaluate the model's ability to transfer reasoning between different dimensional contexts. Specifically, we test whether a model trained on 3D data (ShapeStacks) can generalize to a custom 2D dataset, and vice versa.
    \item \textbf{Dynamics generalization}: We evaluate whether Qwen-SFT-RL (trained on ShapeStacks) can generalize spatio-physical knowledge of stacked systems to dynamic physics context. Specifically, we use the support scenario from the Physion~\cite{physion} dataset as evaluation benchmarks.
    \item \textbf{Height generalization}: We investigate if the model can learn and generalize abstract height-related principles, rather than just memorizing specific height patterns. We achieve this by fine-tuning two separate models on exclusively low-stack and high-stack data, and then evaluating their performance on the unseen height category.
\end{itemize}
Three types of generalization are illustrated in Figure \ref{fig_generalization}. In all generalization experiments, the fine-tuned models are based on Qwen2.5-VL-7B. Except for differences related to the training data—such as the content of the fine-tuning set, the step of the final checkpoint and number of epochs—all other hyperparameters remain the same as those used for Qwen-SFT-RL. Details are provided at Appendix D.1.

\subsubsection{Dimension Generalization: Asymmetry Between 3D and 2D}
Although 3D analysis involves an extra spatial dimension, the core physical principle of analyzing the center of mass projection remains the same. We investigate whether a model trained on data from one spatial dimension can generalize its reasoning capability across different dimensions.

For 2D stacked system data, we generate samples using the Box2D physics engine \cite{box2d}, with tower heights ranging from 3 to 6 (as 2D height-2 towers are overly simple), and refer to this dataset as \emph{2d-towers}. The full dataset is split to training set and test set. For more details on data generation, we refer the readers to Appendix D.2.  

We first evaluate Qwen-SFT-RL on the 2d-towers test set to analyze its generalization ability from 3D to 2D. The evaluation result (Table~\ref{tblgeneralization}) shows a gain of 29.0\% and reveals a clear improvement in generalization.

To test the capability in the reverse direction---from 2D to 3D---we fine-tune the model on the 2d-towers training set. The resulting model is denoted as Qwen-2d. We evaluate Qwen-2d on the test sets of both ShapeStacks and 2d-towers. As shown in Table~\ref{tblgeneralization}, Qwen-2d achieves an 80\% boost in in-domain accuracy and a modest 17.4\% gain when generalizing to ShapeStacks, which is lower than the 29.0\% improvement achieved by Qwen-SFT-RL on 2d-towers.

This suggests that while dimensional generalization is possible, it is more effective from 3D to 2D. We hypothesize this is because 3D samples provide richer spatial information, making it easier for a 3D-proficient model to handle simpler 2D cases.

\subsubsection{Dynamics Generalization: Hard to Generalize}
We next explore the generalization from static to dynamic spatio-physical reasoning. Specifically, we evaluate Qwen-SFT-RL (trained on the ShapeStacks dataset) on the \emph{support scenarios} of the \emph{Physion} benchmark~\cite{physion}, where Physion is a dynamic classical mechanics benchmark that contains a variety of physical scenes. In the support scenario, a stacked geometry system is subjected to an external force, requiring models to predict whether the structure remains stable or collapses, and how the collapse unfolds if it does. This experiment aims to assess whether a model that has learned robust static reasoning about physical stability can transfer the understanding to dynamic scenarios.

The evaluation results of Qwen-SFT-RL are shown in Table~\ref{tblgeneralization}. Clearly, the model proficient in static physical reasoning still struggles to generalize to dynamic samples, showing only a 1\% improvement. This indicates that models struggle to transfer their understanding of stacked system to dynamic domains, even though dynamics scenes contain underlying static equilibrium principles. One possible explanation is the large visual discrepancy between static and dynamic samples, which may hinder the model’s ability to integrate its learned reasoning patterns with new visual contexts.

\subsubsection{Height Generalization: Easier From Low to High}
We investigate height generalization, where the core reasoning task---estimating the center of mass---remains the same despite variations in stack height.
We divide the original training set into ShapeStacks-low (heights 2-3) and ShapeStacks-high (heights 4-6). The resulting models, fine-tuned on high and low data, are denoted as Qwen-high and Qwen-low, respectively. For evaluation, we apply the same height-based categorization to the original test set and randomly sample 1,000 examples from both the low and high subsets.

As shown in Table~\ref{tblgeneralization}, Qwen-low achieves stronger in-domain performance and better height generalization than Qwen-high. The generalization from high to low yields only a 7.9\% improvement, while generalization from low to high shows a 17.8\% gain, which even  surpassing the in-domain performance of Qwen-high (a 16.9\% gain). On mixed-height ShapeStacks set, Qwen-low achieves a 29.7\% performance increase, significantly higher than Qwen-high’s 16.1\%.

These findings indicate that VLMs are capable of height generalization in both directions, but are more effective from lower to higher samples. We attribute this asymmetry to the increased visual and physical complexity of high-stack samples, which can lead to challenging, uninformative gradients during training. In contrast, simpler low-stack samples facilitate the learning of correct reasoning strategies, which results in stronger performance and better generalization.

\subsubsection{Performance Drop with Domain Shift}
Our final analysis evaluates the models across all benchmark scenarios. The results are shown in Table~\ref{tblgeneralization}, where performance improvements on the in-domain samples are highlighted in bold.
We observe a clear trend: \emph{the fine-tuned VLM is capable of meaningful generalization to closely related domains, but this capability decays as the domain gap between test samples and train samples widens.}

Specifically, we observe that: (1) Qwen-SFT-RL's significant gains on ShapeStacks height variations diminish on 2D tasks and are nearly absent on the Physion task. (2) Qwen-low's in-domain gains drop noticeably on out-of-domain ShapeStacks height and 2D samples, with almost no improvement on Physion, though it still surpasses the base model. (3) Qwen-high improves significantly on 2D samples, but its gains decrease on other ShapeStacks height variations and disappear entirely on Physion support. (4) Qwen-2d, despite strong in-domain 2D gains, shows limited improvement on any higher-dimensional ShapeStacks samples and only marginal gains on Physion.

In addition, the comparison shows that Qwen-SFT-RL consistently outperforms Qwen-RL across all generalization test sets, indicating that SFT also contributes to improving the generalization ability of the fine-tuned model.



\section{Conclusion}
Our comprehensive analysis reveals a critical disconnect between the apparent capabilities and the actual physical understanding of current open-source VLMs. We first identify their fundamental limitations, including weak performance, human-like cognitive biases, and a lack of deep reasoning indicating that quality is more important than form. We then demonstrate that while mainstream fine-tuning can offer significant in-domain performance gains, this improvement is largely superficial. The models show limited generalization to novel physical scenarios, especially as the domain gap increases, suggesting a reliance on statistical shortcuts rather than robust physical principles. This exposes a fundamental limitation of current paradigms: they excel at pattern matching but fail to instill the generalizable understanding required for world models. 

Our findings strongly suggest that the path forward lies not in merely scaling existing methods, but in developing new approaches. Future work should prioritize strategies, such as leveraging physically-grounded simulation data and designing novel training paradigms, that explicitly encourage the learning of causal and generalizable physical laws.

\bibliography{aaai2026}

\end{document}